\documentclass{article}

\usepackage{PRIMEarxiv}

\usepackage{amsmath}

\usepackage[utf8]{inputenc} 
\usepackage[T1]{fontenc}    
\usepackage{hyperref}       
\usepackage{url}            
\usepackage{booktabs}       
\usepackage{amsfonts}       
\usepackage{nicefrac}       
\usepackage{microtype}      
\usepackage{fancyhdr}       
\usepackage{graphicx}       
\graphicspath{{media/}}     

\title{Head-tail Loss: A simple function for Oriented Object Detection and Anchor-free models
}

\author{
  Pau Gallés, Xi Chen \\
  Dpt. Cartography and GIS \\
  ECNU - East China Normal University \\
  Shanghai\\
  \texttt{\{pgalles\}@stu.ecnu.edu.cn} \\
}

\begin{document}
\maketitle

\begin{abstract}
This paper presents a new loss function for the prediction of oriented bounding boxes, named head-tail-loss. The loss function consists in minimizing the distance between the prediction and the annotation of two key points that are representing the annotation of the object. The first point is the center point and the second is the head of the object. However, for the second point, the minimum distance between the prediction and either the head or tail of the groundtruth is used. On this way, either prediction is valid (with the head pointing to the tail or the tail pointing to the head). At the end the importance is to detect the direction of the object but not its heading. The new loss function has been evaluated on the DOTA and HRSC2016 datasets and has shown potential for elongated objects such as ships and also for other types of objects with different shapes.
\end{abstract}


\section{Introduction}

Object detection on satellite images is a rapidly growing field, with a wide range of applications such as monitoring land use, natural resource management, and disaster response. The ability to automatically detect and locate objects in satellite images can greatly improve the efficiency and accuracy of many tasks that rely on this data.

Object detection can be classified into two variants: detection using horizontal bounding boxes (HBBs) and detection using oriented bounding boxes (OBBs). HBBs are the traditional approach where objects are detected using rectangular boxes that align with the image axes. On the other hand, OBBs are oriented boxes that can align with the true orientation of the objects in the image. OBBs have been shown to be more accurate than HBBs, as they take into account the rotation of the objects, which is an important aspect in satellite images where objects may be oriented in any direction. Additionally, OBBs are less sensitive to small changes in object orientation, which can lead to more robust object detection. In recent years, several works \cite{li2019dsfd, zhou2019objects, zhou2019bottom} have proposed to use OBBs for object detection in satellite images. These methods have shown promising results and have demonstrated the advantages of using OBBs over HBBs. However, the use of OBBs also brings new challenges, such as the need for more complex models and the increased computational cost. Therefore, the development of efficient and accurate OBB-based object detection methods for satellite images is an important research area that has yet to be fully explored. One example of OBB-based detector is FCOSR which is an object detection model that utilizes a complex loss function to improve performance. Unlike traditional object detection models, such as Faster R-CNN \cite{ren2015faster}, FCOSR uses a multi-task loss function that simultaneously optimizes for both classification and localization. This loss function is designed to handle rotated bounding boxes and is composed of several components, including the centerness-aware loss and the rotated IoU loss \cite{tian2019fcos}. This complex loss function allows FCOSR to better handle rotated objects, leading to improved performance on datasets with a high degree of rotation.

Another important consideration in object detection on satellite images is the difference between anchor-based and anchor-free models. Anchor-based models rely on predefined anchor boxes, whereas anchor-free models do not use anchor boxes and instead directly predict the object's bounding box coordinates. Anchor-free models have been shown to be more efficient and have fewer parameters than anchor-based models, but they may be less robust to variations in object scale. Furthermore, anchor-based models are more sensitive to the choice of anchor sizes and aspect ratios, while anchor-free models are more robust in this regard.

In addition, object detection models can be classified as single-stage or multistage. Single-stage models, such as YOLO \cite{redmon2018yolov3} and FCOS \cite{tian2019fcos}, directly predict the object's bounding box coordinates and class scores in one pass. Multistage models, such as RetinaNet \cite{lin2017focal} and FPN \cite{lin2017feature}, use a two-stage process where a region proposal network is used to generate potential object locations, which are then passed to a second stage for detection and classification. Multistage models tend to be more accurate than single-stage models, but they are also more computationally expensive.

In terms of performance evaluation, several metrics are commonly used for object detection on satellite images, such as precision, recall, Intersection over Union (IoU), mean average precision (mAP), and confusion matrix. Precision measures the proportion of true positive detections among all positive detections, recall measures the proportion of true positive detections among all actual objects, IoU measures the overlap between predicted and ground-truth bounding boxes, mAP is the mean of average precision across all classes and confusion matrix provides the count of true positives, true negatives, false positives, and false negatives. These metrics allow to evaluate the performance of the model by giving a comprehensive understanding of how well the model is detecting objects in the image.

Recently, several state-of-the-art loss functions for oriented object detection have been proposed and used in the literature. Some examples include the polar ray loss \cite{zhang2019polar}, which is a differentiable function based on the angle between the predicted bounding box and the ground-truth bounding box, and has shown promising results in object detection on satellite images. The Rotate IoU loss \cite{liao2020rotate} is a differentiable function based on the intersection over union (IoU) between the predicted bounding box and the ground-truth bounding box. The Rotation-Invariant and Scale-Invariant Loss (RIS-Loss) \cite{zhang2020ris} is a loss function based on the angle-sensitive IoU and the scale-sensitive IoU. Another example is the Centerness-Aware Scale-Adaptive Loss (CASA) \cite{yu2020centerness}, which is a loss function that considers both the centerness and the scale-adaptiveness of the detection results. Other loss functions include the Scale-Aware and Rotation-Aware Loss (SRA-Loss) \cite{zhang2021scale}, the Rotation-Invariant and Scale-Invariant Loss (RIS-Loss) \cite{zhang2020ris}, and the Rotate IoU-NMS loss \cite{ding2020rotate}.

In this paper, we propose a new loss function, named head-tail-loss, for the prediction of oriented bounding boxes in object detection on satellite images, specifically for elongated objects such as ships. Our loss function consists of minimizing the distance between the prediction and the annotation of two key points that represent the annotation of the object: the center point and one of the two extremities of the object, either the head or the tail. By using this method, either the prediction with the head pointing to the tail or the tail pointing to the head is valid, therefore allowing the model to detect the direction of the object, but not its heading. The head-tail-loss is simpler than the existing loss functions and in this paper, we show its feasibility and potential for improving the detection of elongated objects in satellite images.

\section{Methodology}

In this paper, we used two publicly available datasets for the task of object detection on satellite images: DOTA \cite{xia2018dota} and HRSC2016 \cite{liu2016ship}. In this section, we will describe the format and content of these datasets in detail.

DOTA, which stands for "Dataset for Object Detection in Aerial Images," is a large-scale dataset for object detection in aerial images. It contains over 2,800 images and 15 object categories, including plane, ship, storage tank, baseball diamond, tennis court, basketball court, ground track field, harbor, bridge, large vehicle, small vehicle, helicopter, roundabout, soccer-ball field, and swimming pool. Each image in the dataset is annotated with oriented bounding boxes and object categories. The images in the DOTA dataset were collected from Google Earth, and they have a resolution range of about 0.3m to 3m per pixel. The objects in the images vary greatly in scale and orientation, making the dataset challenging for object detection.

HRSC2016, which stands for "High-Resolution Ship Detection Challenge 2016," is a dataset specifically designed for ship detection in high-resolution satellite images. It contains over 1,500 images and one object category (ship) and each image in the dataset is annotated with oriented bounding boxes. The images in the HRSC2016 dataset have a resolution range from 2-m to 0.4-m and the size of images ranges from 300x300 to 1500x900 and most of them are larger than 1000x600. The objects in the images vary greatly in scale and orientation, making the dataset challenging for object detection.

Both DOTA and HRSC2016 datasets provide a challenging testbed for object detection in satellite images due to the large-scale variation and dense cluttered background of the objects in images. In addition, DOTA dataset provides a more diverse set of object classes and the HRSC2016 dataset provides lower-resolution images which allow evaluation of the performance of the model in different scenarios.

The new loss function, named head-tail-loss, is defined as follows:

\begin{equation} \label{eq:1}
L_{head-tail} = \frac{\left\| \boldsymbol{p}_{center} - \boldsymbol{g}_{center} \right\|^2 + \min \left( \left\| \boldsymbol{p}_{head} - \boldsymbol{g}_{head} \right\|^2, \left\| \boldsymbol{p}_{head} - \boldsymbol{g}_{tail} \right\|^2 \right)}{S}
\end{equation}

Where $\boldsymbol{p}{center}$, $\boldsymbol{p}{head}$ are the center and head predictions, respectively and $\boldsymbol{g}{center}$, $\boldsymbol{g}{head}$, $\boldsymbol{g}_{tail}$ are the center, head and tail groundtruth, respectively, S is the image size in pixels. This loss function minimizes the distance between the prediction and the annotation of two key points that represent the annotation of the object: the center point and one of the two extremities of the object, either the head or the tail. The minimum is only between the distances of the predicted head to annotated head and the predicted head to annotated tail. Then the center point distances are added to this minimum, and the final result is divided by the image size with the aim to make the result equal to a small value in the range around 0 and 1. This allows the model to detect the direction of the object, but not its heading, and also makes the loss function independent of the image size.

\begin{figure}
\centering
\includegraphics[width=10.5 cm]{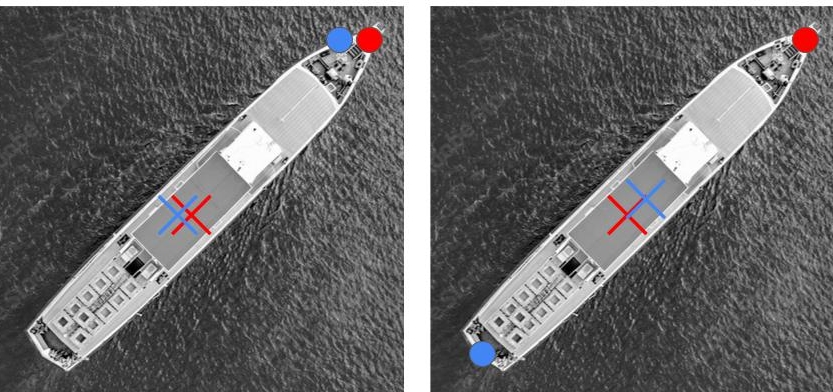}
\caption{Conceptual example of the head-tail-loss function. The image shows two satellite photos of a ship side by side. On the left, the head-tail ground truth or annotation is similar to the prediction. In the right photo, the prediction points to the tail instead, but the prediction is equally valid. The head-tail loss function allows for predictions that point in either direction, as long as the direction of the object is accurately captured, regardless of the heading of the object. This allows the model to detect the direction of the object, but not its heading.
\label{figheadtail}}
\end{figure}

As depicted in Figure~\ref{figheadtail}, the head-tail loss function allows for predictions that point in either direction, as long as the direction of the object is accurately captured, regardless of the heading of the object. This allows the model to detect the direction of the object, but not its heading, and also make the loss function independent of the image size.

Equation \ref{eq:1} does not take into account the width of the object. Initially, the idea was to set the width as a function of the height, specifically for the purpose of detecting ships and vessels. However, it was later decided to add an additional component to the equation that considers the absolute difference between the predicted and the actual width, divided by the size of the image. This modification broadens the applicability of the function to a wider range of object shapes.

\section{Results}

\begin{table}[hbt!]
\caption{Performance results of different models in terms of Mean Average Precision (mAP) and Average Recall (AR).}
\centering
\scriptsize
\begin{tabular}{ccc}
\toprule
\textbf{Model} & \textbf{mAP} & \textbf{AR} \\
\midrule
FCOS+HT & 0.659 & 0.778 \\
FCOS & 0.760 & 0.837 \\
FCOS+HTc & \textbf{0.778} & \textbf{0.857} \\
\bottomrule
\end{tabular}
\label{table:maphrsc}
\end{table}

The performance of the different models trained with the dataset HRSC2016 is presented and analyzed in Table \ref{table:maphrsc}. The first model, $FCOS+HT$, is the experiment using FCOS model and the proposed head-tail loss function. The second model, FCOS+HTc, refers to the same model but keeping the centerness component in the FCOS model. The third model, $FCOS$, is the standard FCOS model. As can be seen in the table, the proposed head-tail loss function with the centerness component removed ($FCOS+HT$) achieved similar performance in terms of mAP and AR compared to the standard FCOS model ($FCOS$) and a slight improvement in terms of mAP when the centerness component is kept ($FCOS+HTc$).

The models trained on HRSC2016 varied considerably depending on the defined random seed. Thus, the results are the average of several experiments\footnote{These random seeds used were: $129$,$305$,$531$,$8$,$98$,$25$,$727$}.

\begin{table}[hbt!]
\caption{Performance results of different object detection models on the DOTA dataset. The scores for each model are expressed as mean average precision (mAP) and average recall (AR) as expressed in the methodology section. The columns represent the different object categories.\label{tab1}}
\centering
\tiny
\begin{tabular}{lccccccccccccccc}
\toprule
\textbf{model} & \textbf{BC} & \textbf{BD} & \textbf{Bridge} & \textbf{GTF} & \textbf{HC} & \textbf{Harbor} & \textbf{LV} & \textbf{Plane} & \textbf{RA} & \textbf{SBF} & \textbf{SP} & \textbf{ST} & \textbf{SV} & \textbf{Ship} & \textbf{mAP} \\
\midrule
FCOSR & 0.840 & 0.798 & 0.485 & 0.707 & \textbf{0.648} & 0.726 & \textbf{0.776} & \textbf{0.895} & 0.652 & \textbf{0.578} & \textbf{0.751} & \textbf{0.863} & \textbf{0.811} & \textbf{0.881} & \textbf{0.754} \\
roitrans & \textbf{0.860} & \textbf{0.846} & \textbf{0.510} & \textbf{0.754} & 0.564 & \textbf{0.734} & 0.762 & 0.894 & 0.625 & 0.616 & 0.707 & 0.845 & 0.744 & 0.871 & 0.749 \\
fasterRCNN & 0.849 & 0.762 & 0.402 & 0.671 & 0.596 & 0.458 & 0.558 & 0.892 & \textbf{0.677} & 0.552 & 0.680 & 0.832 & 0.735 & 0.759 & 0.689 \\
\bottomrule
\end{tabular}
\label{table:mapdotacomplex}
\end{table}

\begin{table}[hbt!]
\caption{Performance results on the DOTA10 dataset using the FCOS model with different variations. The scores are expressed as mean average precision (mAP) and average recall (AR) as explained in the methodology section. \label{tab2}}
\centering
\tiny
\begin{tabular}{lcccccccccccccccc}
\toprule
\textbf{model} & \textbf{BC} & \textbf{BD} & \textbf{Bridge} & \textbf{GTF} & \textbf{HC} & \textbf{Harbor} & \textbf{LV} & \textbf{Plane} & \textbf{RA} & \textbf{SBF} & \textbf{SP} & \textbf{ST} & \textbf{SV} & \textbf{Ship} & \textbf{TC} & \textbf{mAP} \\
\midrule
FCOS+HT & 0.500 & 0.648 & 0.220 & 0.527 & 0.259 & 0.300 & 0.352 & 0.796 & 0.544 & 0.316 & 0.571 & 0.674 & 0.472 & 0.522 & 0.897 & 0.506 \\
FCOS+HTc & 0.749 & 0.779 & 0.362 & 0.581 & 0.494 & 0.492 & 0.691 & 0.880 & 0.593 & \textbf{0.529} & 0.664 & 0.764 & 0.733 & 0.766 & 0.908 & 0.666 \\
FCOS & \textbf{0.807} & \textbf{0.789} & \textbf{0.448} & \textbf{0.589} & \textbf{0.506} & \textbf{0.610} & \textbf{0.739} & \textbf{0.885} & \textbf{0.650} & 0.502 & \textbf{0.693} & \textbf{0.830} & \textbf{0.778} & \textbf{0.844} & \textbf{0.909} & \textbf{0.705} \\
\bottomrule
\end{tabular}
\label{table:mapdotasimple}
\end{table}

The same notation is used for the models that were trained with the dataset DOTA10. Additionally, the two-stage detectors of roitrans\cite{ding2018learning} and fasterRCNN\cite{ren2015faster} as well as FCOSR\cite{tian2019fcosr}, which is a single-stage anchor-free detector for rotated objects, were trained. The results for these alternative models are summarized in Table \ref{table:mapdotacomplex}. And the results with FCOS with the head-tail loss and the original loss are shown in Table \ref{table:mapdotasimple}.

These numbers are clearly worse than in the case of HRSC2016. This second dataset is more difficult and it has objects in different shapes. The head-tail loss function was designed for elongated objects and this is one possible reason for the worse scores.

An interesting finding was that the centerness branch in the FCOS architecture appeared to be unnecessary when using the head-tail loss function, as it already incorporates the center point of the object. However, the results showed that maintaining the centerness branch in the implementation with the head-tail loss function resulted in improved performance in both experiments.

\section{Conclusions}

The new loss function, named head-tail-loss, has been evaluated on the DOTA and HRSC2016 datasets. The results indicated that the head-tail-loss performed better on the HRSC2016 dataset, which contains images of ships, than on the DOTA dataset. This is likely because the head-tail-loss is more suitable for elongated objects such as ships, as observed in the improved performance for the elongated object categories in DOTA. Additionally, an experiment was conducted to evaluate the performance of the head-tail-loss on objects with different shapes, such as storage tanks, by measuring the minimum distance between the predicted head and the annotated head, tail, left side, or right side. The results showed that this approach worked better for squared objects such as storage tanks.

However, the results in the second experiment appear to be worse than expected. One possible explanation is that the models used in this second experiment are more complex, due to the use of multiple stages and the sophistication of the loss function in the case of FCOSR. Additionally, the DOTA10 dataset is more challenging than HRSC2016, which may have also contributed to the worse performance. Therefore, it is important to keep in mind the characteristics of the dataset and the type of objects that the model will be used to detect when selecting a loss function and model architecture.

\section*{Acknowledgments}

The authors would like to acknowledge the valuable guidance provided by Dr. Xi Chen, a researcher in the field of Cartography and Geographical Information Systems. Dr. Chen, who is currently a professor at East China Normal University, is known for his expertise in image processing, machine learning, and remote sensing. His knowledge and insights have been instrumental in the development of this research. Additionally, the authors would like to thank the DOTA and HRSC2016 datasets for providing the data used in this study.

\bibliographystyle{unsrt}  
\bibliography{references}

\end{document}